\documentclass[sigconf]{acmart}
\AtBeginDocument{%
  \providecommand\BibTeX{{%
    \normalfont B\kern-0.5em{\scshape i\kern-0.25em b}\kern-0.8em\TeX}}}

\usepackage{comment}

\begin{document}
\title{Automatic diagnosis of knee osteoarthritis severity using Swin transformer}


\author{ Aymen Sekhri }
 \email{asekhri@inttic.dz}
 \affiliation{%
   \institution{Laboratoire PRISME, université d'Orléans}
   \streetaddress{12 Rue de Blois}
   \city{Orléans}
   \country{France}
   \postcode{45100}
 }

  \author{Marouane Tliba}
 \email{marouane.tliba@univ-orleans.fr}
 \affiliation{%
   \institution{Laboratoire PRISME, université d'Orléans}
   \streetaddress{12 Rue de Blois}
   \city{Orléans}
   \country{France}
   \postcode{45100}
 }

 \author{ Mohamed Amine Kerkouri}
 \email{mohamed-amine.kerkouri@univ-orleans.fr}
 \affiliation{%
   \institution{Laboratoire PRISME, université d'Orléans}
   \streetaddress{12 Rue de Blois}
   \city{Orléans}
   \country{France}
   \postcode{45100}
 }
 \author{Yassine Nasser }
 \email{yassine.nasser@univ-orleans.fr}
 \affiliation{%
   \institution{Laboratoire PRISME, université d'Orléans}
   \streetaddress{12 Rue de Blois}
   \city{Orléans}
   \country{France}
   \postcode{45100}
 }

 \author{Aladine Chetouani}
 \email{aladine.chetouani@univ-orleans.fr}
 \affiliation{%
   \institution{Laboratoire PRISME, université d'Orléans}
   \streetaddress{12 Rue de Blois}
   \city{Orléans}
   \country{France}
   \postcode{45067}
 }

 \author{Alessandro Bruno}
 \email{alessandro.bruno@iulm.it}
 \affiliation{%
   \institution{IULM AI Lab, IULM University}
   \streetaddress{Via Carlo Bo 1}
   \city{Milan}
   \country{Italy}
   \postcode{20143}
 }

  \author{Rachid Jennane}
 \email{rachid.jennane@univ-orleans.fr}
 \affiliation{%
   \institution{IDP laboratory, université d'Orléans}
   \streetaddress{xxx}
   \city{Orleans}
   \country{France}
   \postcode{45067}
 }

\renewcommand{\shortauthors}{SEKHRI et al.}

\begin{abstract}

Knee osteoarthritis (KOA) is a widespread condition that can cause chronic pain and stiffness in the knee joint. Early detection and diagnosis are crucial for successful clinical intervention and management to prevent severe complications, such as loss of mobility. In this paper, we propose an automated approach that employs the Swin Transformer to predict the severity of KOA. Our model uses publicly available radiographic datasets with Kellgren and Lawrence scores to enable early detection and severity assessment. To improve the accuracy of our model, we employ a multi-prediction head architecture that utilizes multi-layer perceptron classifiers. Additionally, we introduce a novel training approach that reduces the data drift between multiple datasets to ensure the generalization ability of the model. The results of our experiments demonstrate the effectiveness and feasibility of our approach in predicting KOA severity accurately. 

\end{abstract}




\maketitle

\section{Introduction}
\label{sec:intro}
Knee osteoarthritis (KOA) is a degenerative disease of the knee joint and the most common form of arthritis. It affects almost half of the population aged 65 years or older worldwide, causing pain, mobility limitation, and impaired quality of life. KOA is caused by a breakdown of knee articular cartilage and bone micro-architecture changes \cite{kohn2016classifications}. Joint space narrowing, osteophyte formation, and sclerosis are KOA's most visually relevant pathological features that can be visualized with radiographs. Although various imaging techniques such as magnetic resonance, computed tomography, and ultrasound have been introduced to diagnose osteoarthritis, radiography remains the most widely used method for initial diagnosis due to its accessibility, low cost, and widespread use.

\begin{figure}[]
    \centering
    \includegraphics[width=\columnwidth]{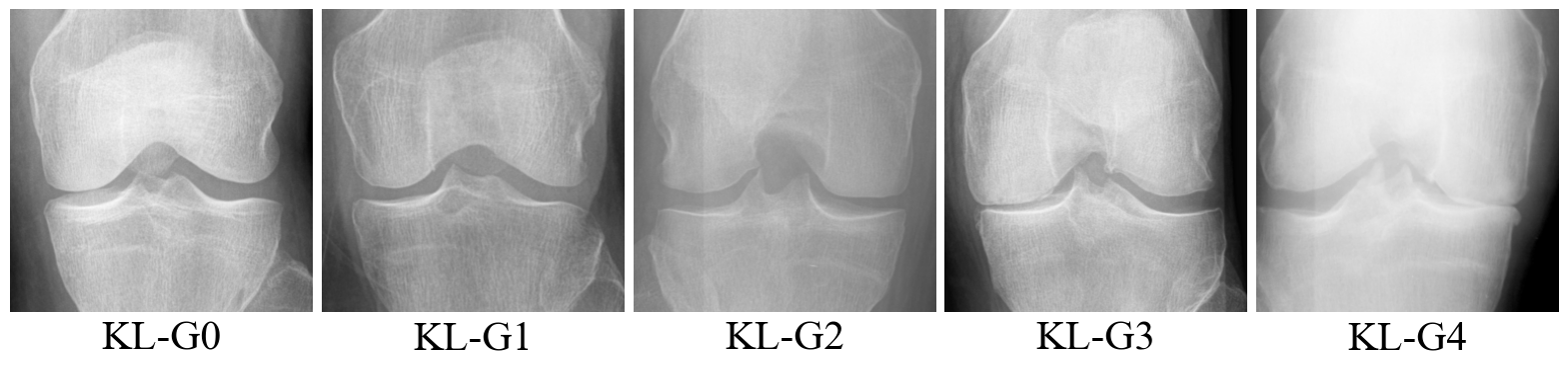}
    \caption{Samples of knee radiographs. KL-G0: healthy knee without osteoarthritis, KL-G1: doubtful osteoarthritis, KL-G2: minimal osteoarthritis, KL-G3: moderate osteoarthritis, and KL-G4: severe osteoarthritis.}
    \vspace{-7mm}
    \label{fig:grades}
\end{figure}

Kellgren and Lawrence (KL) classified KOA severity into five stages based on the radiographic features, from KL-G0 for healthy cases to KL-G4 for severe cases \cite{kohn2016classifications} (See Fig \ref{fig:grades}). However, KOA changes gradually, so the evaluation into different stages is often subjective and depends on the operator. This causes subjectivity and makes the automatic KOA diagnosis a difficult task. In addition, the high similarity between the X-ray images increases the challenge of achieving an accurate diagnosis.

Several deep learning-based methods have been proposed for medical imaging applications \cite{DAQA}, and many to diagnose KOA in recent years. In \cite{antony2017automatic}, Antony \textit{et al.} employed Convolutional Neural Networks (CNNs) to quantify the severity of KOA from radiographic images. Their method is based on two main steps: first, automatically locate the knee joints using a Fully Convolutional Neural etwork (FCN), then, classify the knee joint images using a second CNN. In addition, to improve the quantification of KOA, they combined the classification loss with the regression loss to consider the continuous aspect of the disease progression. 
Tuilpin \textit{et al.} \cite{tiulpin2018automatic} presented a Siamese CNN network for KL grade prediction. They used three models with different random seeds and combined their outputs with a softmax layer to obtain the final KL grade.
Chen \textit{et al.} \cite{chen2019fully} proposed an ordinal loss for fine-tuning various CNN models to classify KOA severity. They leveraged the ordinal nature of the knee KL grading system and penalized incorrect classifications more by increasing the distance between the real and predicted KL grades.
Nasser et al. \cite{nasser2020discriminative} proposed a Discriminative Regularized Auto-Encoder (DRAE) for early KOA prediction using X-ray images. The proposed model uses a discriminative penalty term and the traditional AE reconstruction cost function to enhance the separability of the features learned from different classes. The aim was to boost the recognition system's performance by minimizing the inter-class variance and maximizing the intra-class distance. 
Recently, transformers have shown promising results in various medical imaging tasks \cite{shamshad2023transformers}. Wang \textit{et al.} \cite{wang2023transformer} proposed a novel data augmentation method for early detection of KOA using a Vision Transformer model. The method involves shuffling the position embedding of non-ROI patches and exchanging the ROI patches with other images. The authors also used a hybrid loss function that combines label smoothing and cross-entropy to improve the model's generalization capability and avoid over-fitting. 
Several important studies \cite{brahim2019decision},\cite{gatti2018neuralseg}, \cite{riad2018texture}, \cite{swiecicki2021deep}, \cite{antony2017automatic}, used two multi-center databases, the Osteoarthritis Initiative (OAI, \url{https://nda.nih.gov/oai/}) and the Multicenter Osteoarthritis Study (MOST, \url{https://most.ucsf.edu/}) by not accounting for the data drift problem. The latter occurs when a machine learning model trained on one dataset lowers its performance when tested on another set of data. Subsequently, data drift causes poor generalization and performance degradation. 



In this work, we first investigate the use of the Swin transformer in predicting KOA severity from radiographic images. In particular, the Swin transformer is the core network that extracts high-level features and detects KOA-induced changes. Second, we introduce a multi-predictive classification header to address the high similarity problem between different KOA grades. In addition, to reduce the data drift problems between the data in the two databases, OAI and MOST, we tested several learning strategies to find the one providing the model with better generalization capabilities and balanced classification results.

The remainder of the paper is organized as follows: the proposed method is described in Section \ref{sec:method}. Next, the obtained experimental results are presented in Section \ref{sec:Exp}. Finally, the conclusions and outlooks are given in Section \ref{sec:conclusion}.

\section{Proposed Method}
\label{sec:method}
\begin{figure}[]
    \centering
    \includegraphics[ scale = 0.75]{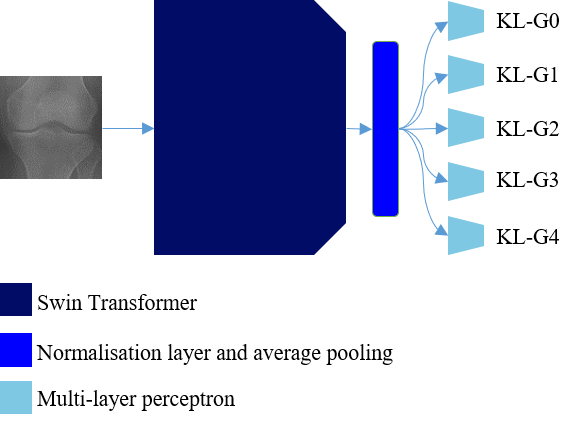}
    \setlength{\belowcaptionskip}{-3mm}
    \caption{Swin transformer architecture with a multiple prediction head architecture}
    \label{fig:arch}
\end{figure}

The method proposed in this paper consists of two parts: 1) a Swin transformer as a features extractor and 2) a multi-prediction head network as a classifier. The schematic illustration of our proposed network is presented in Figure \ref{fig:arch}.


\subsection{Swin Transformer}
\label{subsec:Swin}

The Swin Transformer  \cite{liu2021swin} is a state-of-the-art model that has been specifically designed to address the challenges of applying transformer models in the visual domain. While transformers have been widely successful in natural language processing, they have been less effective in computer vision due to the unique characteristics of visual data. The Swin Transformer proposes a novel architecture that leverages hierarchical feature maps and shift-based windows to improve the efficiency and performance of the model. With its innovative approach, the Swin Transformer has emerged as one of the most efficient and effective transformer models for visual applications. The model is divided into four stages, where the features are hierarchically extracted in each stage.

The input image with dimensions $H \times W \times 3$ is divided into $\frac{H}{4} \times \frac{W}{4}$  non-overlapping patches as tokens of size $4\times 4 \times 3 = 48$. 
These tokens are then passed through the first stage, consisting of a linear embedding layer and two Swin Transformer blocks. The linear embedding layer projects the tokens into a higher-dimensional space denoted by $C$; after that, in the first Swin Transformer block, the multi-headed window self-attention mechanism (W-MSA) is employed. This mechanism computes self-attention only between patches within the same window, where each window contains $M\times M$ patches. The second Swin Transformer block utilizes shifted window multi-headed self-attention (SW-MSA), in which the partitioning windows are shifted by $ (\lfloor\frac{M}{2} \rfloor, \lfloor\frac{M}{2} \rfloor)$ patches with respect to the standard partitioning windows used in the previous block. This approach aims to create more relationships between neighboring patches previously located in different windows and reduce the computational complexity of the global MSA module used in vision transformer. 


In the second stage, a patch merging layer is applied to group each $2\times 2$ neighboring patches into a single patch of length $4C$, thus reducing the number of patches to $\frac{H}{8} \times \frac{W}{8}$. These patches are then linearly projected to a dimension of size $2C$ and passed to two Swin Transformer blocks as in the first stage.

This process is repeated in the third stage, using 18 Swin Transformer blocks to produce $\frac{H}{16} \times \frac{W}{16}$ patches of length $4C$. Finally, in the fourth stage, two Swin Transformer blocks are used to produce $\frac{H}{32} \times \frac{W}{32}$ of length $8C$. These consecutive stages jointly produced a hierarchical representation like those of typical convolutional networks.

\subsection{Multi-Prediction Head Network}
\label{subsec:multihead}

The main task of our designed model is to be able to predict the KOA severity grade. This presents a case of a multi-class classification task. Traditionally this is solved by using a single MLP classification head with $5$ outputs activated by a softmax function. 

The complex nature of X-ray images imposes a high similarity between the images of adjacent KL Grades as shown in Figure \ref{fig:grades}.
To address this issue, we decompose the task into multiple binary classification tasks. We use 5 MLP networks, each specializing in predicting one KL-Grade. This enhances the model's ability to extract and filter a rich representation for each class. 


Let $f: X \rightarrow Z$ be our feature extractor, where $X$ and $Z$ are the input and latent spaces, respectively. $x$ represents the input image and $y$ their corresponding one hot encoding label. The predictive label $\hat{y}_i$ at the head classifier $MLP_i$ is defined as:

\begin{equation}
      \hat{y}_i = MLP_i(f(x))
    \label{eq:dists}
\end{equation}

The final predictive label $\hat{y}$ is computed then as follows: 
\begin{equation}
      \hat{y} = argmax(\bigcup_{i = 0 }^{ 4 }\hat{y}_i)  
    \label{eq:dists}
\end{equation}
where $i \in \{0 \dots 4\} $ represents the KL grades.\\




To sum up, our final model consists of a basic Swin-B encoder with $C=128$ and {$2, 2, 18, 2$} Swin Transformer blocks, followed by Normalisation and average pooling layers to produce a final representation vector of size $1024$. This vector is then passed to $5$ MLPs, one for each KL grade. Each MLP contains 3 linear layers of size $384$, $48$, $48$, $1$, respectively. The final layer of each MLP network has a single neuron to predict the occurrence probability of each grade.


\subsection{Data Drift Correction}
\label{subsec:datashift}

In this paper, we employ 2 of the most widely used datasets for KOA classification (i.e. MOST and OAI datasets). These datasets were collected over a substantial amount of time, from several medical centers, and were annotated by a multitude of medical practitioners. The inherent disparity of equipment, study subjects, radiography, and diagnostics methods between different medical centers caused a shift between the datasets 
as further discussed in Section \ref{subsec:latent}. 

We represent our model using the formula $h = g \circ f$, where $f : X \rightarrow Z$ and $g : Z \rightarrow Y$, represent the feature extractor and the multi-classification head, respectively.  $X$ is the input image, $Z$ is the latent feature space, and $Y$ represents the label space.

To address the issue of data drift between the MOST and OAI datasets, we need to align the latent representational spaces between $Z_{MOST}$ and ${Z}_{OAI}$. This means that the feature extractor $f$ needs to be able to perceive the data distributions from $\mathcal{D_{MOST}}$ and $\mathcal{D_{OAI}}$ as belonging to the same distribution $\mathcal{D}$. It models relevant mutual features while discarding any dataset-specific information that could be considered noisy. This could be represented using the following equation:

\begin{equation}
      \mathcal D = ( \mathcal{D_{MOST}} \cup  \mathcal{D_{OAI}} ) \smallsetminus ( \mathcal N_{MOST} \cup \mathcal N_{OAI} )    
    \label{eq:dists}
\end{equation}
where $\mathcal N_{MOST}$ and $\mathcal N_{OAI}$ represent the noisy distribution of information specific to the MOST and OAI datasets, respectively.\\ 


To achieve this result, we train the model $h$ on the MOST dataset and then freeze the MLP layers $g$. We continue to train the feature extractor $f$ on the OAI dataset. This way, we force the feature extractor $f$ to align the representational space for both datasets. This proposed approach leverages the pre-trained source model effectively and adapts it to the target dataset by minimizing the shift between the data distributions in the latent representational space $Z$. The objective is to achieve this without compromising the prior knowledge of the pre-trained classifier.




\subsection{Implementation}
\label{subsec:implemnt}

In order to train the model, we used the AdamW optimizer \cite{loshchilov2017decoupled} with a learning rate of $3e-5$, a weight decay of $0.05$, an epsilon of $1e-8$, and betas of ($0.9, 0.999$) to adjust the weights. We trained the model with a batch size of 32 images for 300 epochs. We implemented the code in PyTorch and used an NVIDIA RTX A4000 GPU with 16 GB of VRAM to speed up the training process. 

We also implemented various data augmentation techniques such as 15-degree rotation, translation, scaling, random horizontal flipping, and contrast adjustment with a factor of 0.3. These techniques have previously been used in similar studies to improve the performance of deep learning models on image classification tasks in order to address the problem of limited data and overfitting.

\section{Experimental Results}
\label{sec:Exp}

To evaluate the efficacy of the proposed approach, we conducted five experiments, described in this section.

\vspace{-1mm}
\subsection{Datasets}
\label{subsec:dataset}
In this study, we employed two widely used and publicly available datasets: 

\textbf{MOST dataset: } It contains 18,269 knee images that were segmented in the same manner as in \cite{tiulpin2018automatic}. We divided this dataset into three subsets, namely training, validation, and testing with a ratio of 6:1:3. Table \ref{tab:most} provides a summary of the dataset's partitioning. We use this dataset to train and evaluate our model's performance on knee image classification.

\textbf{OAI dataset: } It consists of 8260 already prepared knee images \cite{chen2019fully}. It is randomly divided into three subsets, namely training, validation, and testing with a ratio of 7:1:2. Table \ref{tab:oai} summarizes the partitioning of the OAI dataset. We use this dataset to validate and test our model's performance.
\begin{table}[h!]
\small
\centering
\begin{tabular}{ l | c c c c c c}
\hline
& KL-G0 & KL-G1 & KL-G2 & KL-G3 & KL-G4 & \textbf{Total}\\
\hline
Training & 4380 & 1759 & 1827 & 1986 & 1008 & \textbf{10960}\\ 
\hline
Validation & 730 & 294 & 304 & 331 & 168 & \textbf{1827}\\ 
\hline
Testing & 2190 & 880 & 914 & 994 & 504 & \textbf{5482}\\
\hline
\end{tabular}
\caption{\label{tab:most}Label distribution of the MOST dataset}
\vspace{-7mm}
\end{table}

\begin{table}[h!]
\small
\centering
\begin{tabular}{ l | c c c c c c}
\hline
& KL-G0 & KL-G1 & KL-G2 & KL-G3 & KL-G4 & \textbf{Total}\\
\hline
Training & 2286 & 1046 & 1516 & 757 & 173 & \textbf{5778}\\ 
\hline
Validation & 328 & 153 & 212 & 106 & 27 & \textbf{826}\\ 
\hline
Testing & 639 & 296 & 447 & 223 & 51 & \textbf{1656}\\
\hline
\end{tabular}
\caption{\label{tab:oai}Label distribution of the OAI dataset}
\vspace{-9mm}
\end{table}

\subsection{Experimental Protocol}
\label{subsec:protocol}

During the development of our model, we tested multiple configurations and compared them. 
In the first experiment, we use a single classifier to predict all grades simultaneously. In the second experiment, we use the same settings but employed the Multi-prediction head architecture, which involves breaking down the multi-classification problem into sub-binary classifications. For experiments three and four, we explored the data drift between two datasets by training only one dataset per experiment. Finally, in the fifth experiment, we tackled the issue of data drift by transferring the knowledge from the trained classifier on the source dataset (MOST) and solely training the feature extractor of our model on the target dataset (OAI).


\vspace{-1mm}
\subsection{Quantitative Evaluation}
\label{subsec:quant}

\begin{figure}[h!]
    \centering
    \includegraphics[width=\columnwidth]{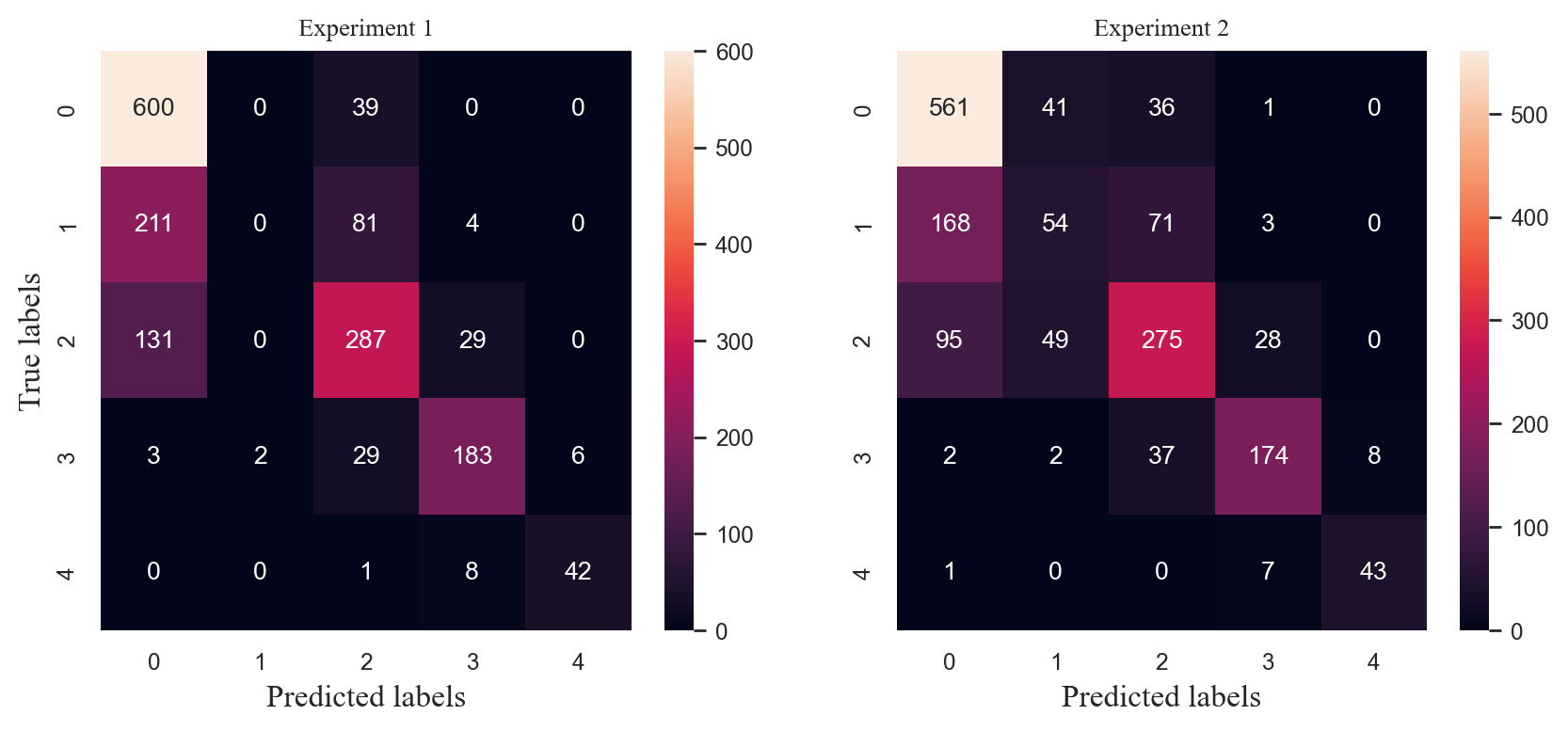}
    \caption{\label{figure:cm}Confusion matrices on the OAI test set of Experiment 1 and 2.}
    \vspace{-3mm}
\end{figure}

\begin{figure*}[!ht]
    \centering
    \includegraphics[height= 35mm]{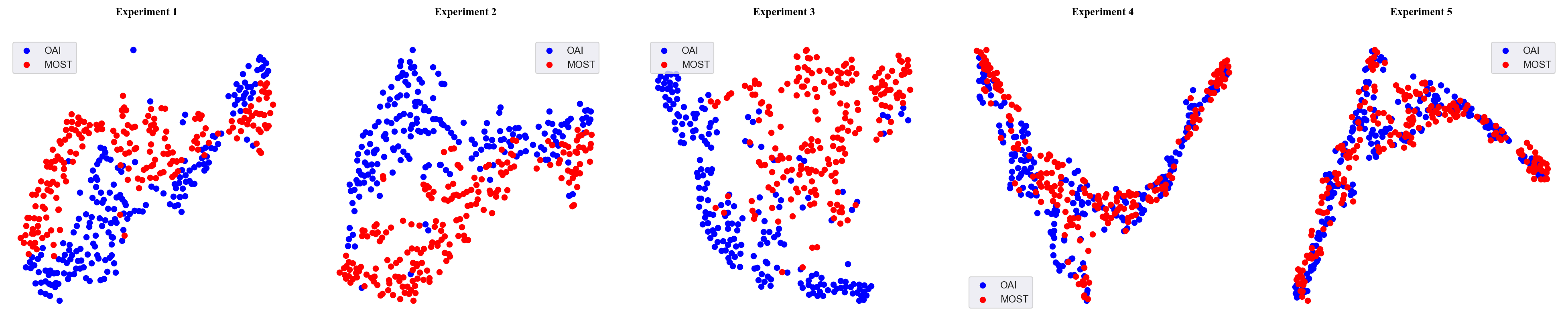}
    \caption{\label{figure:data_shift}t-SNE visualizations of features learned by the model in each dataset.}
\end{figure*}

\begin{figure*}[!ht]
    \centering
    \includegraphics[width= \linewidth]{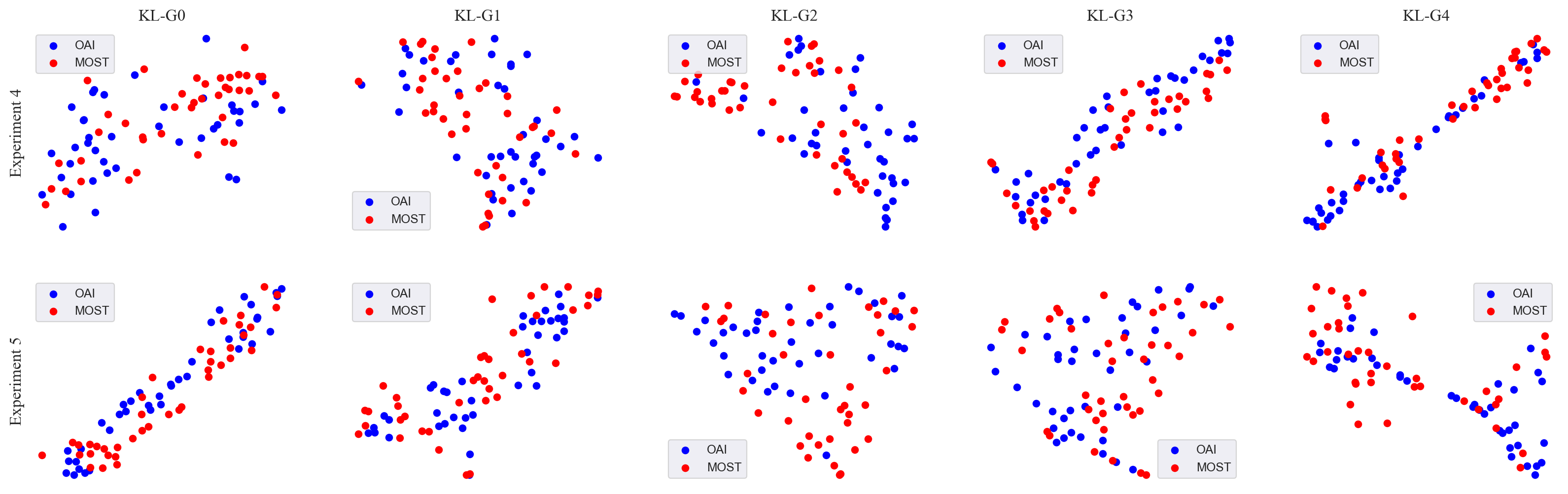}
    \caption{\label{figure:data_shift2}t-SNE visualizations of features learned by the model in experiments 4 and 5 for each dataset.}
    \vspace{+2mm}
\end{figure*}

\begin{figure*}[!ht]
    \centering
    \includegraphics[height= 40mm]{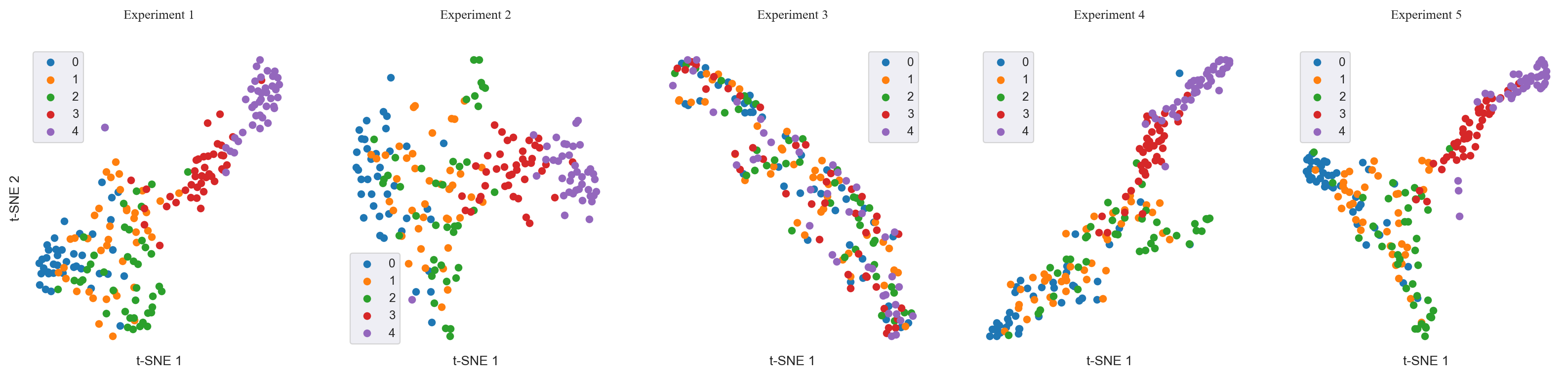}
    \caption{\label{fig:cls_sep}t-SNE visualizations of grades separability in the OAI testset}
\end{figure*}

\begin{table}[htbp!]
\small
\begin{center}
\begin{tabular}{ c c c c c  }
\hline
\textbf{Exp.} & \multicolumn{2}{c}{\textbf{MOST}} & \multicolumn{2}{c}{\textbf{OAI}} \\ 
 & \textbf{Acc (\%) $\uparrow$ } & \textbf{F1 $\uparrow$ } & \textbf{Acc (\%) $\uparrow$ } & \textbf{F1 $\uparrow$ }   \\ 
\hline
 1 &  71.93  &  0.622   & 67.15  & 0.615  \\ \hline 
 2 &  73.13  &  0.684   & 66.85  &  0.657 \\ \hline 
 3 &  39.95  &  0.114   & 38.59  & 0.111  \\ \hline 
 4 &  \textbf{75.43}  &  \textbf{0.714}   & 62.86  &  0.615  \\ \hline 
 5 &  73.25  &  0.667   & \textbf{70.17}  &  \textbf{0.671} \\ \hline 
 
\end{tabular}
\caption{\label{tab:Ablation}Comparison of the five experiments in terms of accuracy and F1 score on the OAI and MOST test sets.}
\end{center}
\vspace{-6mm}
\end{table}

The performances obtained for each considered configurations are presented in Table \ref{tab:Ablation}. In the first two experiments, we observed an improvement in the F1 score for our model when using the Multi-prediction head architecture in the second experiment. Specifically, the model yielded a 0.062 and 0.042 F1 score increase compared to the first experiment in the MOST and OAI test sets, respectively. We also notice an increase in accuracy on the MOST dataset.

Moreover, as seen by the confusion matrices in Figure \ref{figure:cm}, the architecture proposed in experiment 2 was able to avoid the catastrophic failure of detecting the KL-G1 observed in experiment 1. The grad KL-G1 is notoriously challenging to detect even for trained doctors due to the high similarity with the KL-G0 and KL-G2. In fact, the model correctly predicted 54 images in KL-G1 in experiment 2, while 0 images were classified in experiment 1. These results highlight the impact of dividing the multi-classification problem into sub-binary classification problems as described in sections \ref{subsec:multihead}.       
The substantial drop of performance in experiment 3 on both datasets is mainly attributed to the lack of a sufficient quantity of data. Transformer-based models are known to require a lot of data for training \cite{cao2022training}. This has led to the underfitting of our model as it was not able to extract meaningful representations from this dataset. On the other hand, we notice that the performance of the model on the MOST dataset is quite similar, this is due to the richness of the representations in this dataset. 
In experiment 4, the MOST dataset contains more samples that cover a broader range of KOA severity levels than the OAI dataset as shown in Table \ref{tab:most}. Consequently, MOST provides a more diverse and representative training set for our model, leading to better performance in the MOST test set. However, we still see a greater decrease in performance on the OAI dataset compared to experiment 2 in terms of accuracy and F1 score. 
Experiment 5 showed a considerable enhancement in performances on the OAI dataset compared to all other experiments, achieving a $70.17\%$ accuracy and $0.671$ F1-score, as shown in Table \ref{tab:Ablation}, while maintaining a high accuracy on the MOST dataset. This particularly highlights the significance and effectiveness of our method to reduce the data drift and align the latent representations of both datasets as described in section \ref{subsec:latent}.



\subsection{Latent Representation Ability}
\label{subsec:latent}

The reduction of the data drift is an important task for our model as shown in the previous quantitative results. Figure \ref{figure:data_shift} depicts the distribution of latent features extracted for the samples of each dataset across the models produced through our previous experiments. We used the t-SNE algorithm \cite{van2008visualizing} in order to reduce the dimensionality of the features. The data drift in the representation of the two datasets is clearly apparent for both experiments 1 and 2. Even though experiment 2 achieved better results, we still noticed the high disparity of performance between datasets. Due to the underfitting of the model in experiment 3, it was also unable to address the data drift.  In experiment 4 the model was trained only on the MOST dataset. Because of the availability of data, we noticed a better general alignment for data distribution between datasets. But Figure \ref{figure:data_shift2} shows that the shift on the scale of individual classes is still noticeable.  In experiment 5, we noticed a very strong alignment for both datasets on the general and class-specific levels in Figures  \ref{figure:data_shift} and \ref{figure:data_shift2}, respectively. 
Our approach successfully aligned all the data points from both datasets, effectively mitigating the data drift problem. As a result, the learned representations were more relevant to the task, and the model's performance improved significantly.


Figure \ref{fig:cls_sep} illustrates the distribution of latent representations of each class for each of our previous experiments on  the OAI test-set. It highlights the ability of the model  to discriminate and separate the different classes of KL-Grade.  
In experiment 3 where the underfitting occurred, we can observe  the inability of the model to separate the distributions of the different classes.  In experiments 1,2 and 4, the models were able to clearly separate the distributions of KL-G3 and KL-G4. Separating the KL-G0, KL-G1, and KL-G2 grades was more challenging in the first experiment due to the significant similarity between them and the use of a single MLP classifier. Along with the ability to align the distributions of both datasets, we noticed in Experiment 5  a better separability between KL-G0, KL-G1, and KL-G2  which posed a challenge in other experiments. We observed a clear ability to discriminate between KL-G1 and KL-G2 especially, while KL-G0 and KL-G1 still pose some challenges because they represent the none existence and the very early stages of OA respectively.   

Overall, these results demonstrate the effectiveness of our method in handling data drifts and enhancing the model's ability to differentiate between grades of KOA.

\vspace{-4mm}
\subsection{Qualitative Evaluation}
\label{subsec:qual}

\begin{figure}[h!]
    \centering
    \includegraphics[width=\columnwidth]{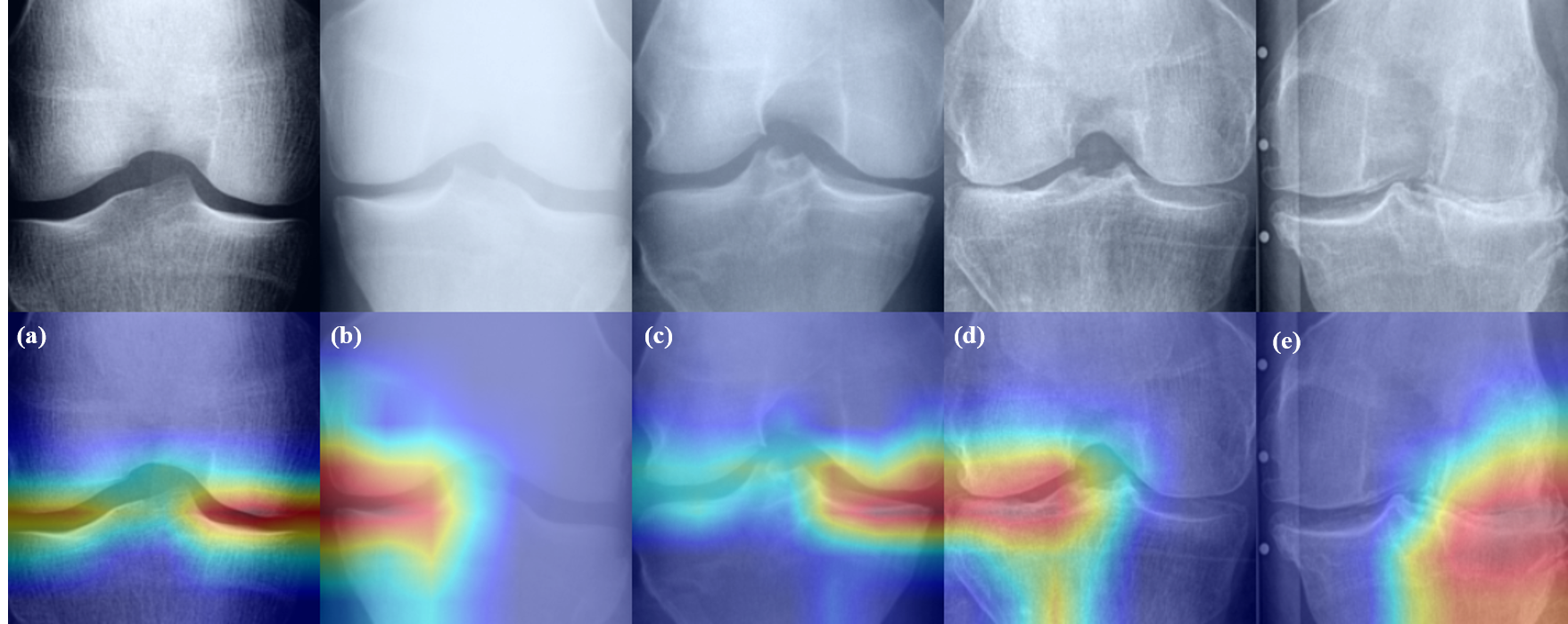}
    \caption{\label{figure:gradCAM_true}GradCAM of the corrected classified images.}
\end{figure}

\begin{figure}[h!]
    \centering
    \includegraphics[width=\columnwidth]{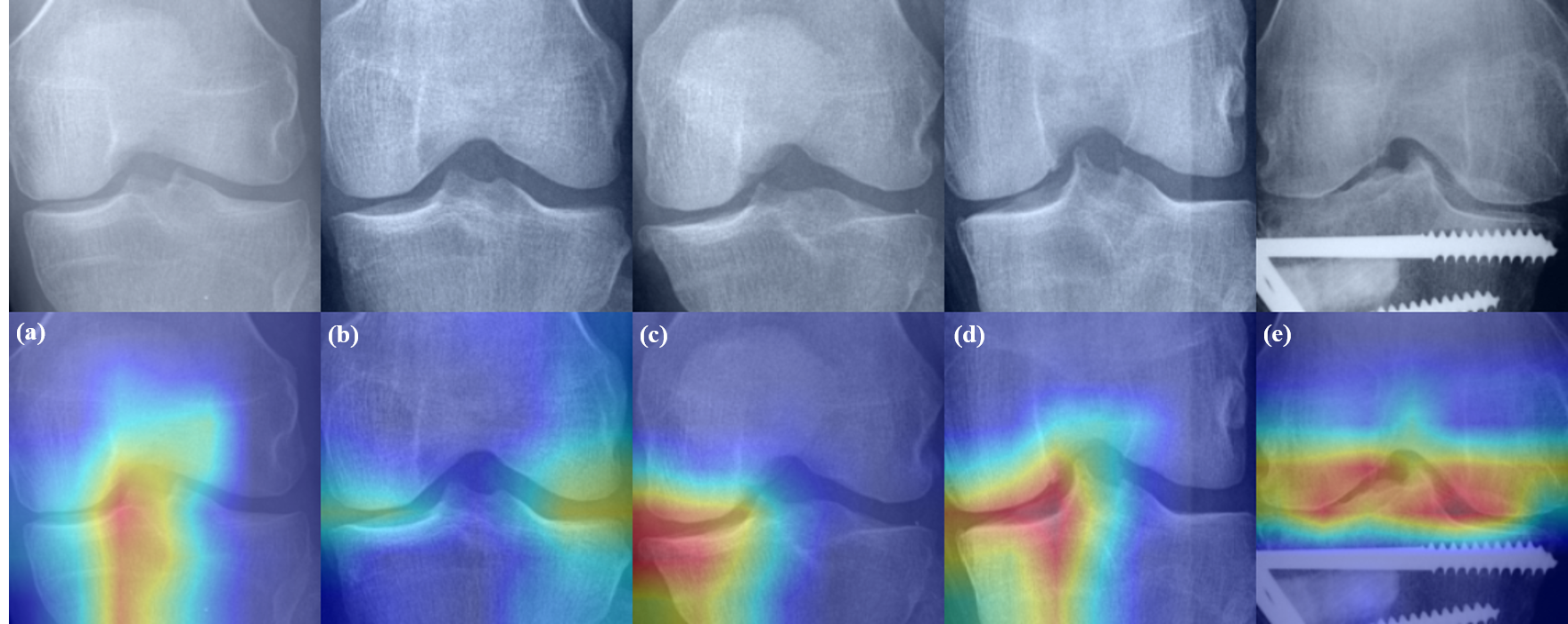}
    \caption{\label{figure:gradCAM_false}GradCAM of the misclassifed images. (a) Predicted KL-G1 (b) Predicted KL-G0 (c) Predicted KL-G1 (d) Predicted KL-G2 (e) Predicted KL-G3.}
\end{figure}

We use GradCAM as a tool for interpretability purposes. By visualizing the last layer's activations of the feature extractor, we chose a sample from each grade, where the true labels of samples from (a) to (e) are from KL-G0 to KL-G4, respectively, as shown in Figures \ref{figure:gradCAM_true} and \ref{figure:gradCAM_false}. 

In Figure \ref{figure:gradCAM_true}, we observed that the model effectively identified areas like osteophytes, joint space narrowing, and sclerosis, which are essential factors for assessing the severity of KOA \cite{lespasio2017knee}.  This points out that our model bases its classifications on the right regions of interest commonly used in clinical diagnosis and not on non-relevant features. 

Figure \ref{figure:gradCAM_false} represents misclassified samples. As can be observed, the model still focuses on the relevant regions around the knee joint. For instance, the model predicts sample (a) as KL-G1, even though the true KL grade was zero. It focused on the area where a medial joint space narrowing was present, which is a possible feature of KL-G1. Similar misclassifications occurred for samples (b), (c), and (d), where the model either overestimated or underestimated the KL grade, indicating the challenge of distinguishing between grades due to their high similarity and also the fact that the KL grade suffers from subjectivity/ambiguity among experts \cite{tiulpin2020automatic}.
In sample (e), we encountered an image that contained an unusual object (i.e. A screw) in the tibia, which could potentially distract the model from the areas of the image that are crucial for grading KOA. However, our model demonstrated robustness by still being able to focus on the region of interest. Furthermore, our model classified the image as a KL-G3 instead of KL-G4, which are close compared to other KL-Grades. This result highlights the ability of our model to prioritize task-specific important features in the image and not be affected by irrelevant and noisy distractors.

\subsection{State-of-the-art Comparison}
\label{subsec:sota}

Table \ref{tab:SOTAComp} presents a comparison of the results obtained with state-of-the-art methods. We note that the methods used in these studies were trained differently. Specifically, some methods used the OAI training set exclusively, others used the MOST training set exclusively, and others used both bases. This diversity in learning can have an impact on the overall performance, and should therefore be carefully considered when interpreting the results. 

Antony et al. \cite{antony2016quantifying} and \cite{antony2017automatic} achieved accuracies of 53.40\% and 63.60\%, respectively, and F1-scores of 0.43 and 0.59, respectively. Chen et al. \cite{chen2019fully} used ordinal loss with different deep learning architectures and achieved accuracies of 69.60\%, 66.20\%, and 65.50\% with Vgg19, ResNet50, and ResNet101, respectively, but they did not report F1-score. Tiulpin et al. \cite{tiulpin2018automatic} used a Siamese network and reported an accuracy of 66.71\%. Wang et al. \cite{wang2021automatic} achieved an accuracy of 69.18\%.

Our proposed method, experiment 5, outperformed all other methods with an accuracy of $70.17\%$ and an F1-score of $0.67$. These results indicate the potential of our proposed method for improving the accuracy and reliability of knee osteoarthritis diagnosis, which could be valuable in clinical practice.


\begin{table}[htbp!]
\small
\begin{center}
\begin{tabular}{ l c c  }
\hline
\textbf{Method} & \textbf{Acc (\%) $\uparrow$ } & \textbf{F1 $\uparrow$ }   \\ 
\hline
 Antony et al. 2016 \cite{antony2016quantifying}  & 53.40 &  0.43 \\ \hline
 Antony et al. 2017 \cite{antony2017automatic}  & 63.60 &  0.59 \\ \hline
 Ordinal Loss (Vgg19) \cite{chen2019fully} & 69.60 &  - \\ \hline
 Ordinal Loss (ResNet50) \cite{chen2019fully}  &  66.20 &   - \\ \hline
 Ordinal Loss (ResNet101) \cite{chen2019fully} & 65.50 &  - \\ \hline
 Siamese net \cite{tiulpin2018automatic}  & 66.71 &  - \\ \hline
 Wang et al. \cite{wang2021automatic}  & 69.18 &  - \\ \hline
 \textbf{Ours experiment 5}  & \textbf{70.17} &  \textbf{0.67} \\ \hline 
\end{tabular}

\caption{\label{tab:SOTAComp} Results for OAI dataset.}
\end{center}
\end{table}
\vspace{-9mm}

\section{Conclusion}
\label{sec:conclusion}
In this paper, we proposed a new method to predict the severity of Knee OA from radiographic images using the Swin Transformer. Our results showed that this method achieved state-of-the-art performance on the OAI test set, significantly outperforming existing methods. We show that the Swin Transformer network is effective in extracting relevant knee OA information, which can be used to detect most of the symptoms of the disease. In addition, handling the data drift and using the multi-prediction head architecture significantly improves the accuracy of the model and helps reduce the similarity between features of nearby grades. Prospects for future work may involve other imaging modalities such as MRI, while exploring clinical and demographic data, to further improve the prediction of KOA severity.

 \begin{acks}
Funded by the TIC-ART project, Regional fund (Region Centre-Val de Loire)
\end{acks}


\bibliographystyle{ACM-Reference-Format}
\bibliography{ref}



\end{document}